\pdfoutput=1

\documentclass[letterpaper, 10 pt, conference]{tex_style/ieeeconf}
\IEEEoverridecommandlockouts
\usepackage{graphics} 
\usepackage{epsfig} 
\usepackage{tabularx}
\usepackage{booktabs}
\usepackage{color}
\usepackage{subfigure}
\usepackage{amssymb}
\usepackage{balance}

\definecolor{CommentRed}{rgb}{0.7,0,0}
\definecolor{CommentGreen}{rgb}{0,0.7,0}
\definecolor{CommentBlue}{rgb}{0,0,0.7}

\title{\LARGE \bf
Vote3Deep: Fast Object Detection in 3D Point Clouds Using Efficient Convolutional Neural Networks
}

\author{Martin Engelcke, Dushyant Rao, Dominic Zeng Wang, Chi Hay Tong, Ingmar Posner
\thanks{Authors are from the Oxford Robotics Institute, University of Oxford.
\texttt{\{firstname\}@robots.ox.ac.uk}}
}

\begin{document}

\maketitle
\thispagestyle{empty}
\pagestyle{empty}

\begin{abstract}

This paper proposes a computationally efficient approach to detecting objects natively in 3D point clouds using convolutional neural networks (CNNs).
In particular, this is achieved by leveraging a feature-centric voting scheme to implement novel convolutional layers which explicitly exploit the sparsity encountered in the input.
To this end, we examine the trade-off between accuracy and speed for different architectures and additionally propose to use an $\mathcal{L}_1$ penalty on the filter activations to further encourage sparsity in the intermediate representations.
To the best of our knowledge, this is the first work to propose sparse convolutional layers and $\mathcal{L}_1$ regularisation for efficient large-scale processing of 3D data.
We demonstrate the efficacy of our approach on the KITTI object detection benchmark and show that \emph{Vote3Deep} models with as few as three layers outperform the previous state of the art in both laser and laser-vision based approaches by margins of up to 40\% while remaining highly competitive in terms of processing time.

\end{abstract}


\section{INTRODUCTION}
3D point cloud data is ubiquitous in mobile robotics applications such as autonomous driving, where efficient and robust object detection is pivotal for planning and decision making. Recently, computer vision has been undergoing a transformation through the use of convolutional neural networks (CNNs) (e.g. \cite{krizhevsky2012imagenet, simonyan2014very, szegedy2015going, he2015deep}).
Methods which process 3D point clouds, however, have not yet experienced a comparable breakthrough.
We attribute this lack of progress to the computational burden introduced by the third spatial dimension.
The resulting increase in the size of the input and intermediate representations renders a naive transfer of CNNs from 2D vision applications to native 3D perception in point clouds infeasible for large-scale applications.
As a result, previous approaches tend to convert the data into a 2D representation first, where nearby features are not necessarily adjacent in the physical 3D space -- requiring models to recover these geometric relationships.

In contrast to image data, however, typical point clouds encountered in mobile robotics are spatially sparse, as most regions are unoccupied.
This fact was exploited in \cite{wang2015voting}, where the authors propose \emph{Vote3D}, a feature-centric \emph{voting algorithm} leveraging the sparsity inherent in these point clouds.
The computational cost is proportional only to the number of \emph{occupied} cells rather than the total number of cells in a 3D grid.
\cite{wang2015voting} proves the equivalence of the voting scheme to a dense convolution operation and demonstrates its effectiveness by discretising point clouds into 3D grids and performing exhaustive 3D sliding window detection with a linear Support Vector Machine (SVM). Consequently, \cite{wang2015voting} achieves the previous state of the art in both performance and processing speed for detecting cars, pedestrians and cyclists in point clouds on the object detection task from the popular KITTI Vision Benchmark Suite \cite{Geiger2012CVPR}.

\begin{figure}[t]
    \parbox[t]{\columnwidth}{
    \centering
    \subfigure[3D point cloud detection with CNNs]{\includegraphics[width=0.95\columnwidth]{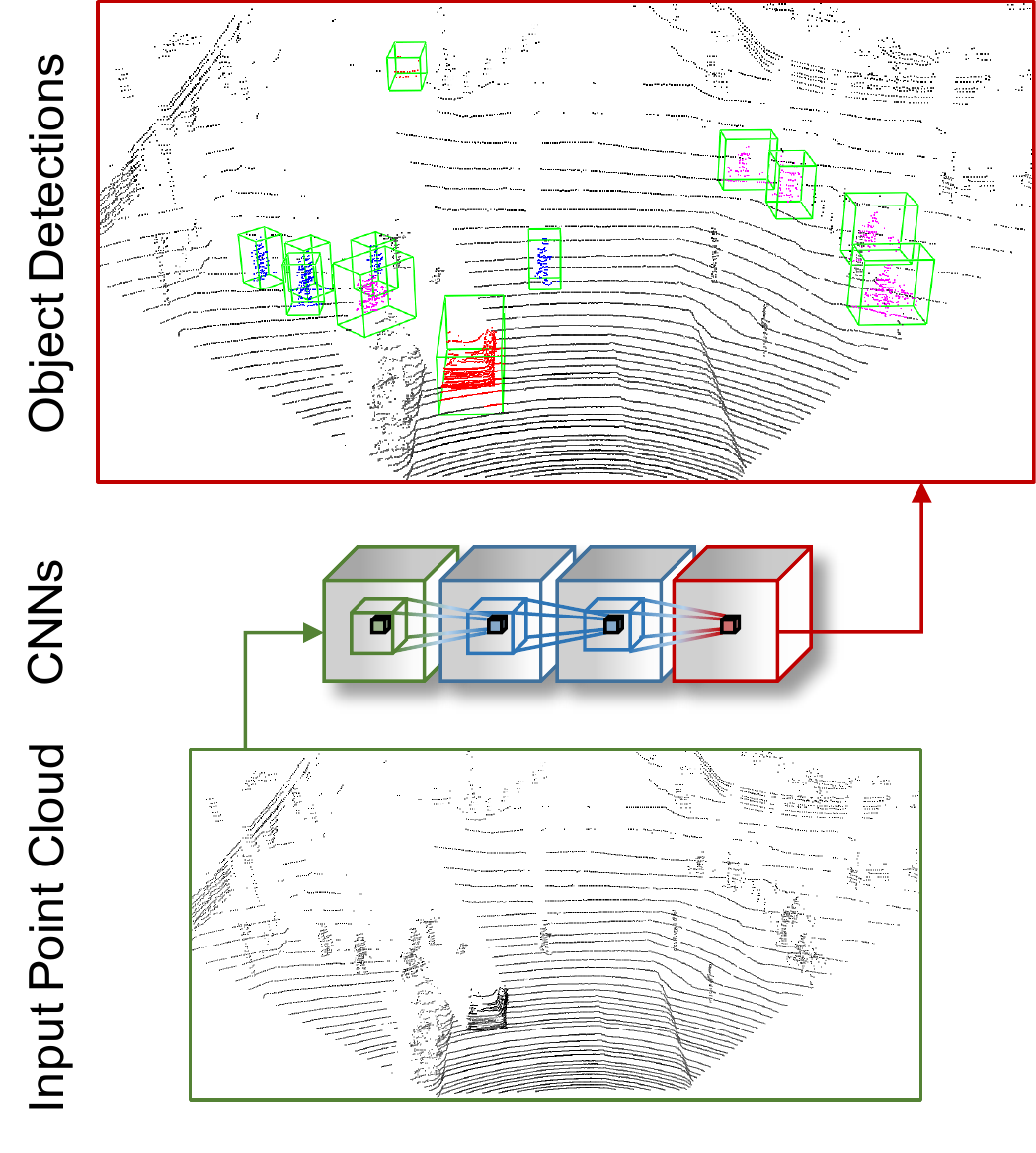}}
    \subfigure[Reference image]{\includegraphics[width=0.95\columnwidth]{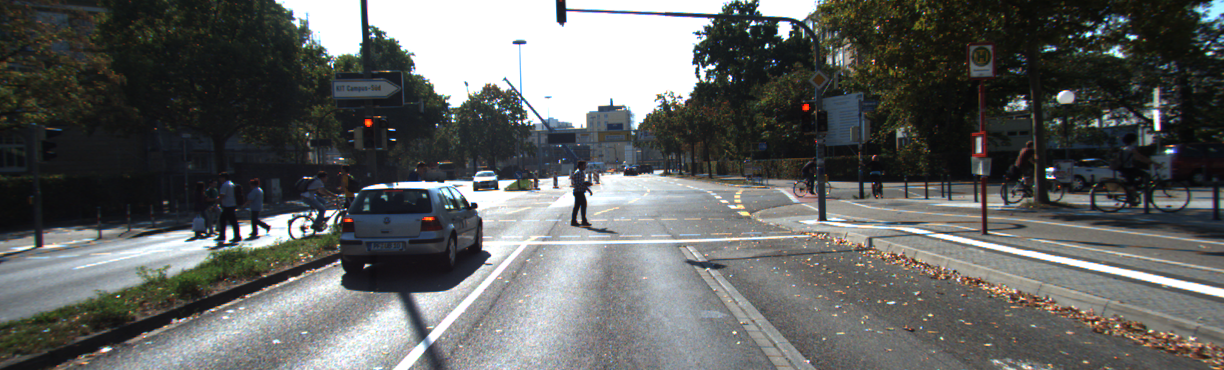}}
    \caption{The result of applying \emph{Vote3Deep} to an unseen point cloud from the KITTI dataset, with the corresponding image for reference. The CNNs apply sparse convolutions natively in 3D via \emph{voting}. The model detects cars (red), pedestrians (blue), and cyclists (magenta), even at long range, and assigns bounding boxes (green) sized by class. Best viewed in colour.}
    \label{fig:pc_detections}}
\end{figure}

Inspired by \cite{wang2015voting}, we propose to exploit feature-centric voting to build efficient CNNs to detect objects in point clouds \emph{natively} in 3D -- that is to say without projecting the input into a lower-dimensional space first or constraining the search space of the detector (Fig. \ref{fig:pc_detections}).
This enables our approach, named \emph{Vote3Deep}, to learn high-capacity, non-linear models while providing constant-time evaluation at test-time, in contrast to non-parametric methods.
Furthermore, in order to enhance the computational benefits associated with sparse inputs throughout the entire CNN stack, we demonstrate the benefits of encouraging sparsity in the inputs to intermediate layers by imposing an $\mathcal{L}_1$ model regulariser during training.

To the best of our knowledge, this is the first work to propose sparse convolutional layers based on voting and $\mathcal{L}_1$ regularisation for efficient processing of full 3D point clouds with CNNs at scale.
In particular, the contributions of this paper can be summarised as follows:
\begin{enumerate}
\item the construction of efficient convolutional layers as basic building blocks for CNN-based point cloud processing by leveraging a voting mechanism to exploit the inherent sparsity in the input data;
\item the use of rectified linear units and an $\mathcal{L}_1$ sparsity penalty to specifically encourage \emph{data} sparsity in the intermediate representations in order to exploit sparse convolutional layers throughout the entire CNN stack.
\end{enumerate}

We demonstrate that \emph{Vote3Deep} models with as few as three layers achieve state-of-the-art performance amongst purely laser-based approaches across all classes considered on the popular KITTI object detection benchmark.
\emph{Vote3Deep} models exceed the previous state of the art in 3D point cloud based object detection in average precision by a margin of up to 40\% while only running slightly slower in terms of detection speed.


\section{RELATED WORK}
\label{sec:related}

A number of works have attempted to apply CNNs in the context of 3D point cloud data.
A CNN-based approach in \cite{livehicle} obtains comparable performance to \cite{wang2015voting} on KITTI for car detection by projecting the point cloud into a 2D depth map, with an additional channel for the height of a point from the ground.
Their model predicts detection scores and regresses to bounding boxes.
However, the projection to a specific viewpoint discards valuable information, which is particularly detrimental, for example, in crowded scenes.
It also requires the network filters to learn local dependencies with regards to depth, information that is readily available in a 3D representation and which can be efficiently extracted with sparse convolutions.

Dense 3D occupancy grids obtained from point clouds are processed with CNNs in \cite{maturana2015voxnet} and \cite{maturana20153d}.
With a minimum cell size of $0.1\mathrm{m}$, \cite{maturana2015voxnet} reports a speed of $6\mathrm{ms}$ on a GPU to classify a single crop with a grid-size of $32 \times 32 \times 32$ cells.
Similarly, a processing time of $5\mathrm{ms}$ per $\mathrm{m}^3$ for landing zone detection is reported in \cite{maturana20153d}.
With 3D point clouds often being larger than $60\mathrm{m} \times 60\mathrm{m} \times 5\mathrm{m}$, this would result in a processing time of $60 \times 60 \times 5 \times 5 \times 10^{-3} = 90\mathrm{s}$ per frame, which does not comply with speed requirements typically encountered in robotics applications.

An alternative approach that takes advantage of sparse representations can be found in \cite{graham2014spatially} and \cite{graham2015sparse}, in which sparse convolutions are applied to comparatively small 2D and 3D crops respectively.
While the convolutional kernels are only applied at sparse feature locations, the presented algorithm still has to consider neighbouring values which take a value of either zero or a constant bias, leading to unnecessary operations and memory consumption.
Another method for performing sparse convolutions is introduced in \cite{jampani2016learning} who make use of ``permutohedral lattices", but only consider comparatively small inputs, as opposed to our work.

CNNs have also been applied to dense 3D data in biomedical image analysis (e.g. \cite{chen2016voxresnet,dou2016automatic,prasoon2013deep}). 
A 3D equivalent of residual networks \cite{he2015deep} is utilised in \cite{chen2016voxresnet} for brain image segmentation.
A cascaded model with two stages is proposed in \cite{dou2016automatic} for detecting cerebral microbleeds.
A combination of three CNNs is suggested in \cite{prasoon2013deep}.
Each CNN processes a different 2D plane and the three streams are joined in the last layer.
These systems run on relatively small inputs and in some cases take more than a minute for processing a single frame with GPU acceleration.



\section{Methods}
\label{sec:approach}

This section describes the application of convolutional neural networks for the prediction of detection scores from sparse 3D input grids of variable sizes.
As the input to the network, a point cloud is discretised into a sparse 3D grid as in \cite{wang2015voting}.
For each cell that contains a non-zero number of points, a feature vector is extracted  based on the statistics of the points in the cell.
The feature vector holds a binary occupancy value, the mean and variance of the reflectance values and three shape factors.
Cells in empty space are not stored which leads to a sparse representation.

We employ the voting scheme from \cite{wang2015voting} to perform a sparse convolution across this native 3D representation, followed by a ReLU non-linearity, which returns a new sparse 3D representation.
This process can be repeated and stacked as in a traditional CNN, with the output layer predicting the detection scores.

Similar to \cite{wang2015voting}, a CNN is applied to a point cloud at N different angular orientations in N parallel threads to handle objects at different orientations at a minimal increase in computation time.
Duplicate detections are pruned with non-maximum suppression (NMS) in 3D space.
NMS in 3D is better able to handle objects that are behind each other as the 3D bounding boxes overlap less than their 2D projections.

Based on the premise that bounding boxes in 3D space are similar in size for object instances of the same class, we assume a fixed-size bounding box for each class, which eliminates the need to regress the size of a bounding box.
We select 3D bounding box dimensions for each class of interest based on the $95^{\mathrm{th}}$ percentile ground truth bounding box size over the training set.

The receptive field of a network should be at least as large as the bounding box of an object, but not excessively large which would waste computation time.
We therefore employ several class-specific networks which can be run in parallel at test time, each with a different total receptive field size depending on the object class.
In principle, it is possible to compute detection scores for multiple classes with a single network; a task left for future work.

\subsection{Sparse Convolutions via Voting}
When running a dense 3D convolution across a discretised point cloud, most of the computation time is wasted as the majority of operations are multiplications by zero.
The additional third spatial dimension makes this process even more computationally expensive compared to 2D convolutions, which form the basis of image-based CNNs.

\begin{figure}[t]
    \centering
    \includegraphics[trim=0cm 15cm 0cm 0cm, clip=true, width=\columnwidth]{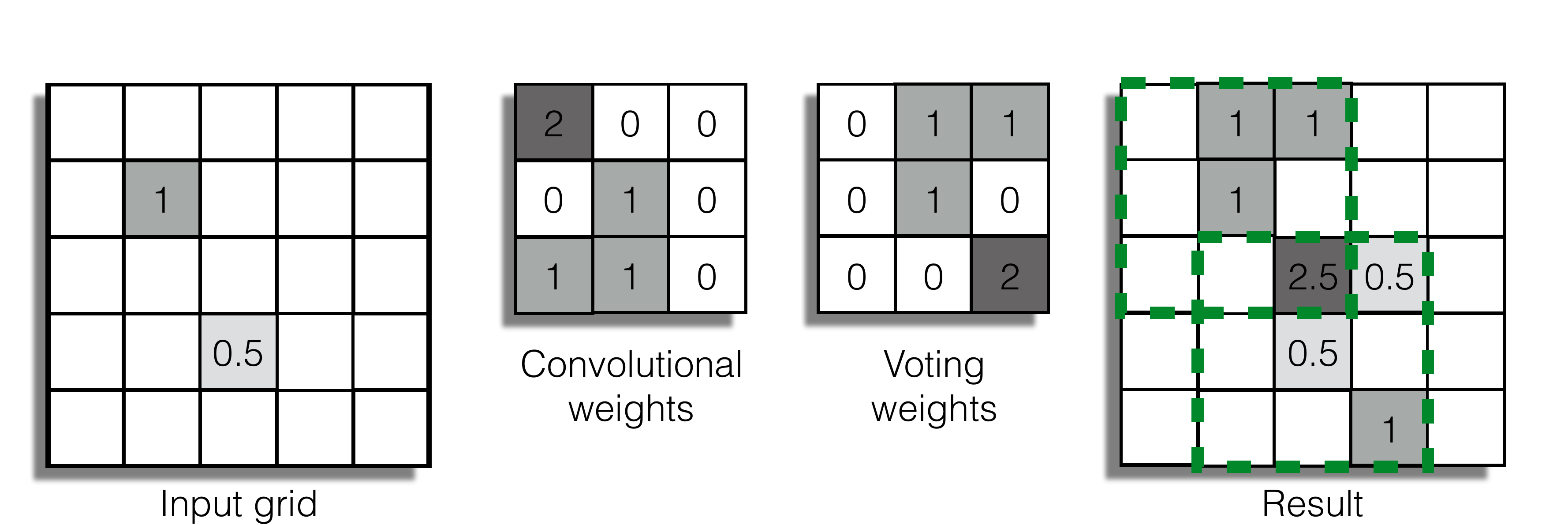}
    \caption{An illustration of the voting procedure on a sparse 2D example input without a bias. The voting weights are obtained by flipping the convolutional weights along each dimension. Whereas a standard convolution applies the filter at every location in the input, the equivalent voting procedure only needs to be applied at each non-zero location to compute the same result. Instead of a 2D grid with a single feature, \emph{Vote3Deep} applies the voting procedure to 3D inputs with several feature maps. For a full mathematical justification, the reader is referred to \cite{wang2015voting}. Best viewed in colour.
    \label{fig:voting}
    }
\end{figure}

Using the insight that meaningful computation only takes place where the 3D features are non-zero, \cite{wang2015voting} introduce a feature-centric voting scheme.
The basis of this algorithm is the idea of letting each non-zero input feature vector cast a set of votes, weighted by the filter weights, to its surrounding cells in the output layer, as defined by the receptive field of the filter.
The voting weights are obtained by flipping the convolutional filter kernel along each spatial dimension. 
The final convolution result is obtained by accumulating the votes falling into each cell of the output (Fig. \ref{fig:voting}).

This procedure can be formally stated as follows.
Without loss of generality, assume we have one 3D convolutional filter with odd-valued kernel dimensions in network layer $c$, operating on a single input feature, with the filter weights denoted by $\mathbf{w}^{c} \in \mathbb{R}^{(2I+1) \times (2J+1) \times (2K+1)}$.
Then, for an input grid $\mathbf{h}^{c-1} \in \mathbb{R}^{L \times M \times N}$, the convolution result at location $(l, m, n)$ is given by:
\begin{equation}
    z_{l, m, n}^{c} = \sum_{i=-I}^{I} \sum_{j=-J}^{J} \sum_{k=-K}^{K} w_{i,j,k}^{c} \; h_{l+i, m+j, n+k}^{c-1} + b^{c}
\end{equation}
\noindent where $b^{c}$ is a bias value applied to all cells in the grid.
This operation needs to be applied to all $L \times M \times N$ locations in the input grid for a regular dense convolution.
In contrast to this, given the set of cell indices for all of the non-zero cells ${\bf\Phi} = \{ \left(l,m,n\right) \; \forall \; h_{l, m, n}^{c-1} \neq 0 \}$, the convolution can be recast as a feature-centric voting operation, with each input cell casting votes to increment the values in neighbouring cell locations according to:
\begin{equation}
    z_{l+i, m+j, n+k}^{c} = z_{l+i, m+j, n+k}^{c} + w_{-i,-j,-k}^{c} \; h_{l, m, n}^{c-1}
\end{equation}
which is repeated for all tuples $ \left(l,m,n\right) \in {\bf\Phi} $ and where $ \{ i, j, k  \in \mathbb{Z} \; | \; i \in \; \left[-I, I\right], j \in \; \left[-J, J\right], k \in \; \left[-K, K\right] \}$.

The voting output is passed through a ReLU non-linearity which discards non-positive features as described in the next subsection.
Crucially, the biases are constrained to be non-positive as a single positive bias would return an output grid in which almost every cell is occupied with a feature vector, hence eliminating sparsity.
The bias $b^{c}$ therefore only needs to be added to each non-empty output cell.

With this sparse voting scheme, the filter only needs to be applied to the occupied cells in the input grid, rather than convolved over the entire grid.
The algorithm is described in more detail in \cite{wang2015voting}, including formal proof that feature-centric voting is equivalent to an exhaustive convolution.

\subsection{Maintaining Sparsity with ReLUs}
The ability to perform fast voting in all layers is predicated on the assumption of sparsity in the input to each individual layer.
While the input point cloud is sparse, the regions of non-empty cells are dilated by each successive convolutional layer, approximately by the receptive field size of the corresponding  filters in the layer.
It is therefore critical to select a non-linear activation function which helps to maintain sparsity in the inputs to each convolutional layer.

This is achieved by applying a rectified linear unit (ReLU) as advocated in \cite{glorot2011deep} after a sparse convolutional layer.
The ReLU activation can be written as:
 \begin{equation}
     h^{c} = \max \left(0, z^{c}\right)
 \end{equation}
 with $z^{c}$ being the input to the ReLU non-linearity in layer $c$ as computed by a sparse convolution, and $h^{c}$ being the output, denoting the \emph{hidden} activations in the subsequent sparse intermediate representation.

In this case, only features that have a value greater than zero will be allowed to cast votes in the next sparse convolution layer.
In addition to enabling a network to learn non-linear function approximations and therefore increasing its representational capacity, ReLUs effectively perform a thresholding operation by discarding negative feature values which helps to maintain sparsity in the intermediate representations.
Lastly, a further advantage of ReLUs compared to other non-linearities is that they are fast to compute.

\section{TRAINING}
\label{sec:training}

Due to the use of fixed-size bounding boxes, networks can be directly trained on 3D crops of positive and negative examples whose dimensions equal the receptive field size specified by the architecture.

Negative training examples are obtained by performing hard negative mining periodically after a fixed number of training epochs.
The class-specific networks are binary classifiers and we choose a linear hinge loss for training due to its maximum margin property.


\subsection{Linear Hinge Loss}
Given an output detection score $\hat{y} \in \mathbb{R}$, a class label $y \in \{ -1, 1 \}$ distinguishing between positive and negative samples, and the parameters of the network denoted as $\theta$, the hinge loss is formulated as:
\begin{equation}
    L \left( \theta \right) = \max \left( 0, 1 - \hat{y} \cdot y \right)
    \label{eq:hinge_loss}
\end{equation}

The loss in Eq. \ref{eq:hinge_loss} is zero for positive samples that score over $1$ and negative samples that score below $-1$.
As such, the hinge loss drives sample scores away from the margin given by the interval $\left[-1,1\right]$.
As with standard CNNs, the $\mathcal{L}_1$ hinge loss can be backpropagated through the network for training.

\begin{table}
    \parbox[t]{\columnwidth}{
    \centering
    \caption{Kernel dimensions in each layer for the architectures used in the model comparison - ``RF" indicates that the receptive field of the output layer depends on the object class}
    \newcolumntype{Y}{>{\centering\arraybackslash}X}
    \begin{tabularx}{0.32\textwidth}{c c c c}
        \toprule
        Model    & Layer 1   & Layer 2   & Layer 3   \\
        \midrule
        A   & $\mathrm{RF_A}$   & - & - \\
        B   & $3\times3\times3$ & $\mathrm{RF_B}$  & - \\
        C   & $5\times5\times5$ & $\mathrm{RF_C}$  & - \\
        D   & $3\times3\times3$ & $3\times3\times3$ & $\mathrm{RF_D}$  \\
        E   & $5\times5\times5$ & $3\times3\times3$ & $\mathrm{RF_E}$  \\
        \bottomrule
    \end{tabularx}
    \label{tab:architectures}}
\end{table}

\begin{figure}
    \centering
    \includegraphics[width=\columnwidth]{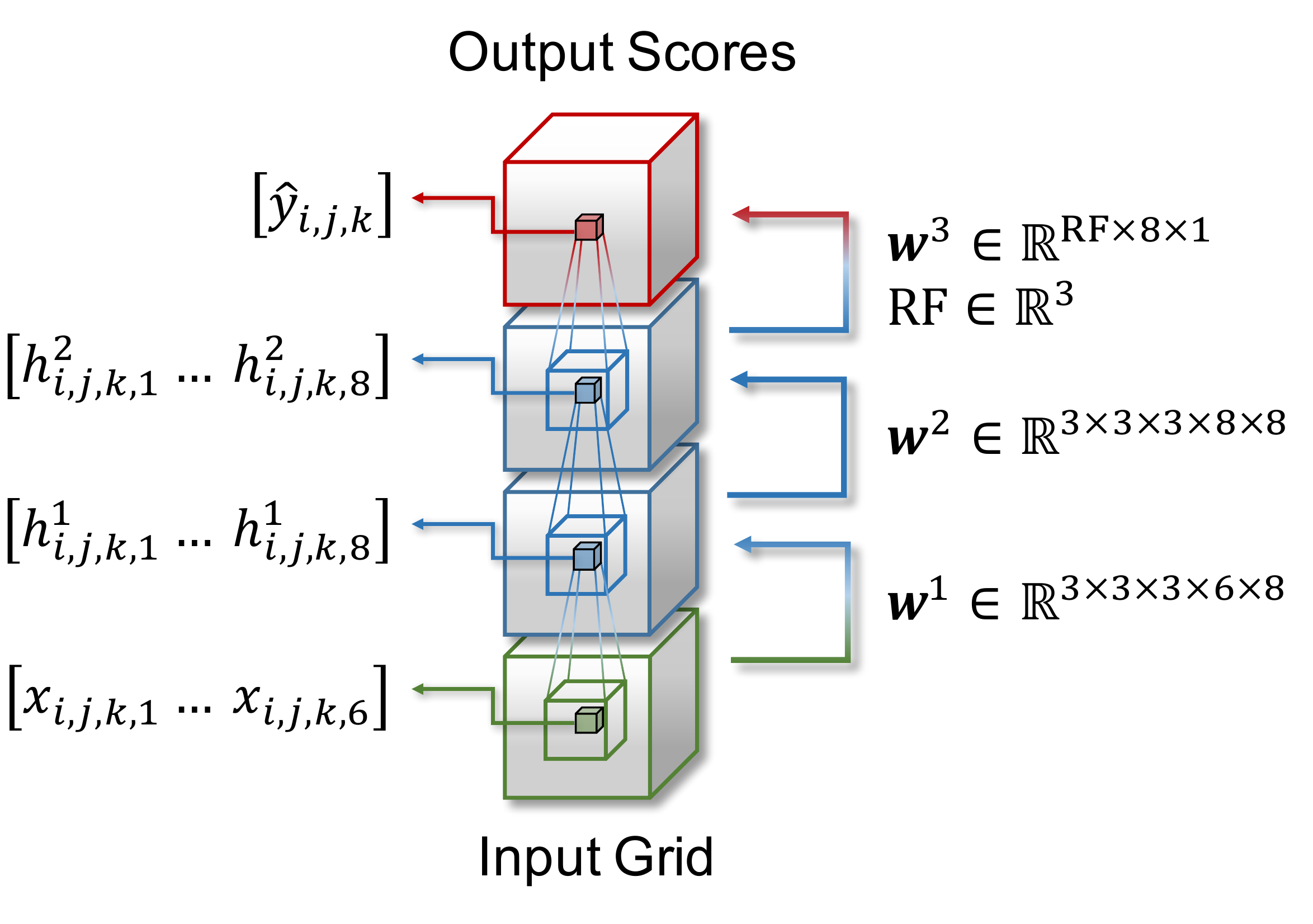}
    \caption{Illustration of the ``Model D'' architecture from Table \ref{tab:architectures}. The input $\bf{x}$ (green) and the intermediate representations $\bf{h}^c$ (blue) for layer $c$ are sparse 3D grids, where each occupied spatial location holds a feature vector (solid cubes). The sparse convolutions with the filter weights $\bf{w}^c$ are performed natively in 3D to compute the predictions (red). Best viewed in colour.}
    \label{fig:architecture_diagram}
\end{figure}


\begin{figure*}
\centerline{
    \subfigure[Cars]{\includegraphics[trim=0cm 6cm 0cm 6cm, clip=true, width=0.27\textwidth]{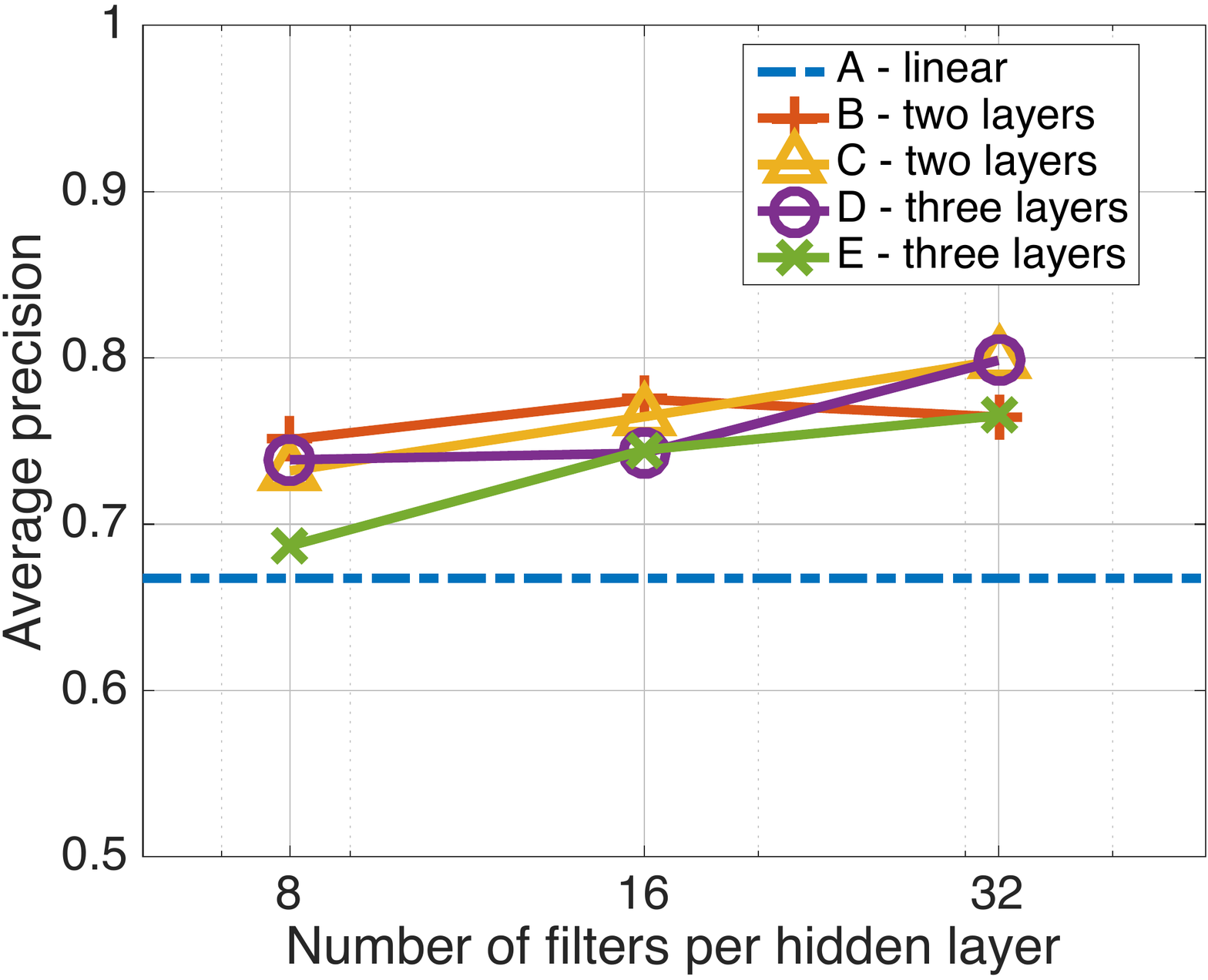}
    \label{fig:model_comparison_car}}
    \hfil
    \subfigure[Pedestrians]{\includegraphics[trim=0cm 6cm 0cm 6cm, clip=true, width=0.27\textwidth]{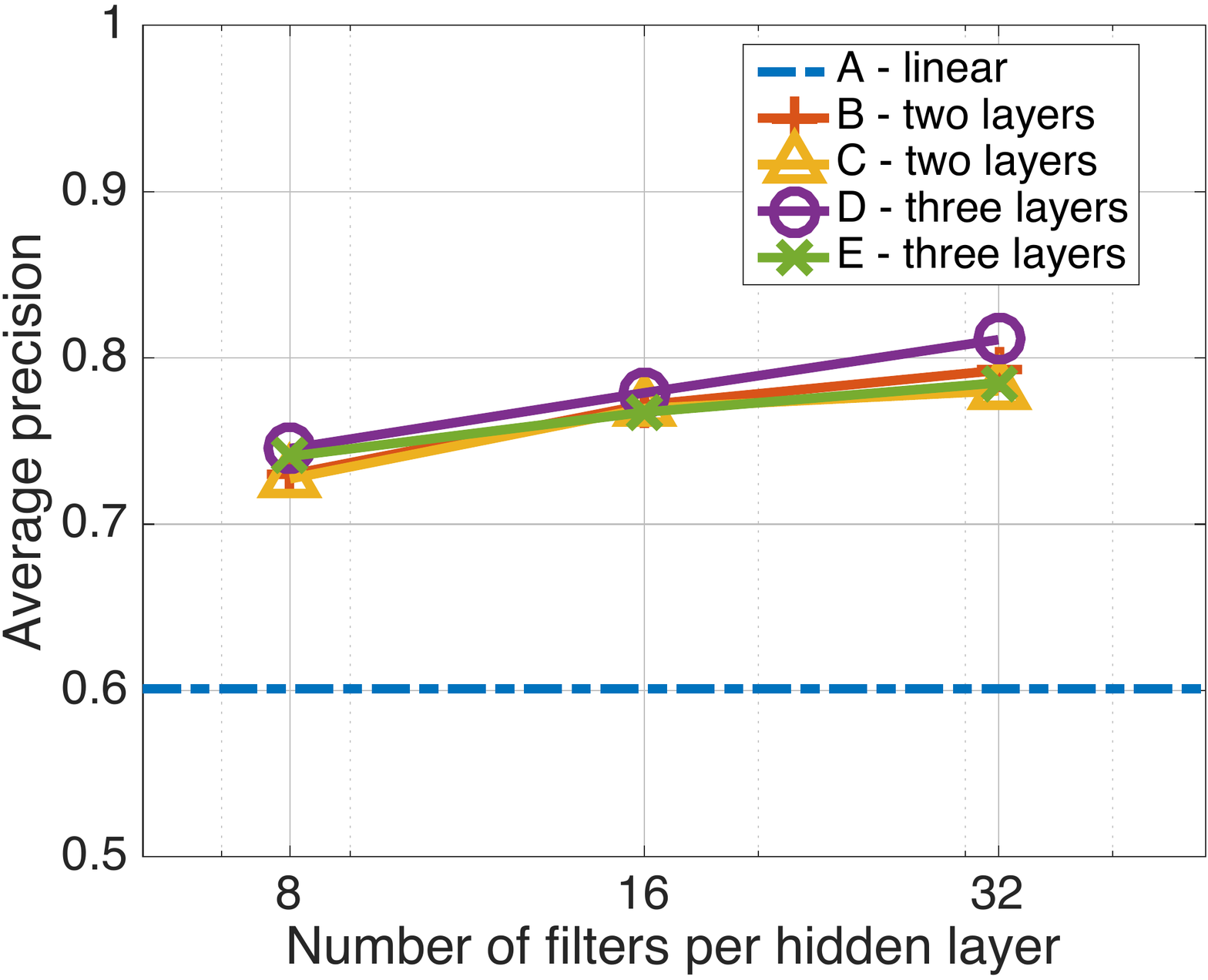}
    \label{fig:model_comparison_ped}}
    \hfil
    \subfigure[Cyclists]{\includegraphics[trim=0cm 6cm 0cm 6cm, clip=true, width=0.27\textwidth]{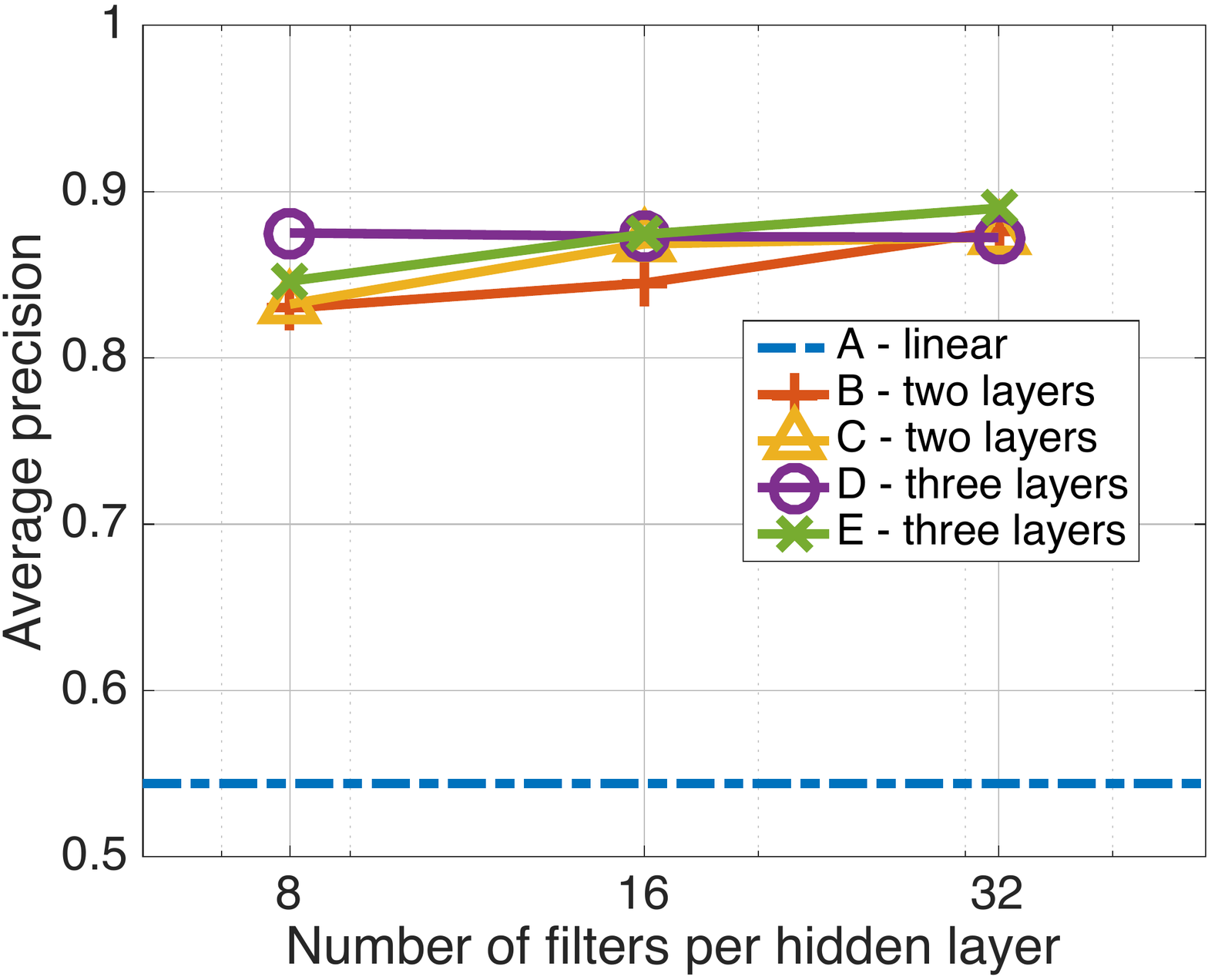}
    \label{fig:model_comparison_cyc}}}
\caption{Model comparison for the architecture in Table \ref{tab:architectures}, showing the average precision for the moderate difficulty level. The non-linear models with two or three layers consistently outperform the linear baseline model our internal \emph{validation set} by a considerable margin for all three classes. The performance continues to improve as the number of filters in the hidden layers is increased, but these gains are incremental compared to the large margin between the linear baseline and the smallest multi-layer models. Best viewed in colour.}
\label{fig:model_comparison}
\end{figure*}

\begin{figure*}
\centerline{
    \subfigure[Cars]{\includegraphics[trim=0cm 6cm 0cm 6cm, clip=true, width=0.27\textwidth]{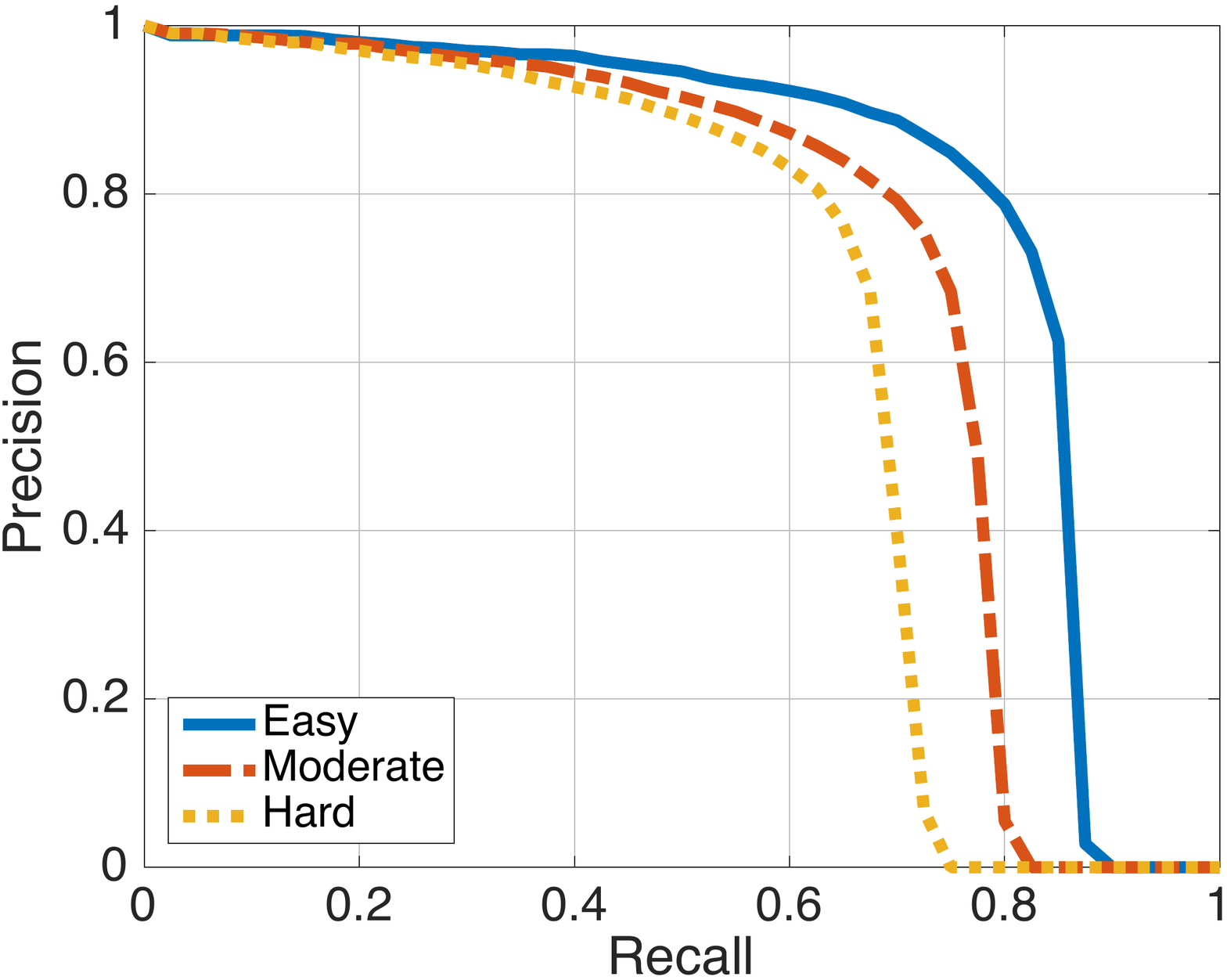}
    \label{fig:kitti_pr_car}}
    \hfil
    \subfigure[Pedestrians]{\includegraphics[trim=0cm 6cm 0cm 6cm, clip=true, width=0.27\textwidth]{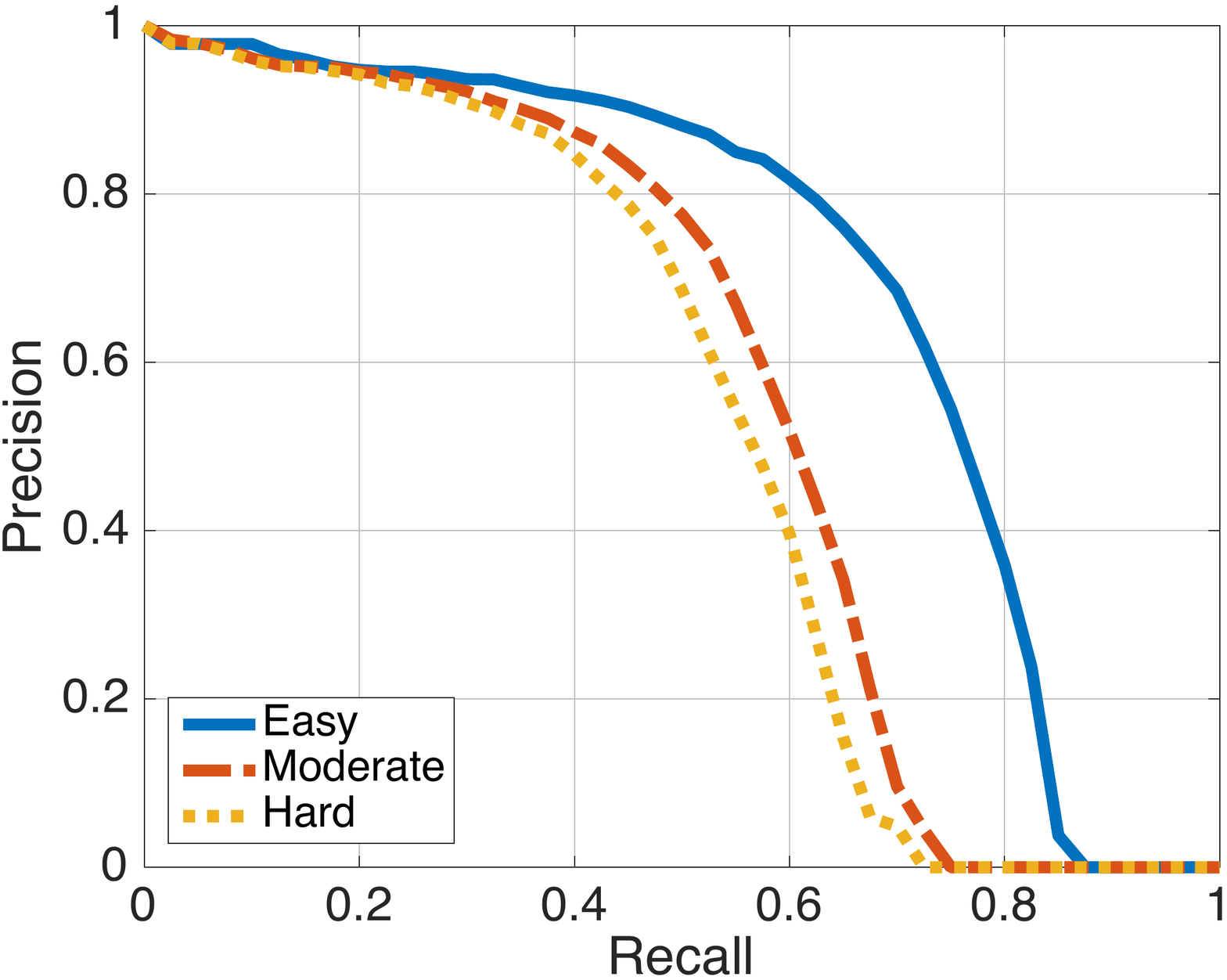}
    \label{fig:kitti_pr_ped}}
    \hfil
    \subfigure[Cyclists]{\includegraphics[trim=0cm 6cm 0cm 6cm, clip=true, width=0.27\textwidth]{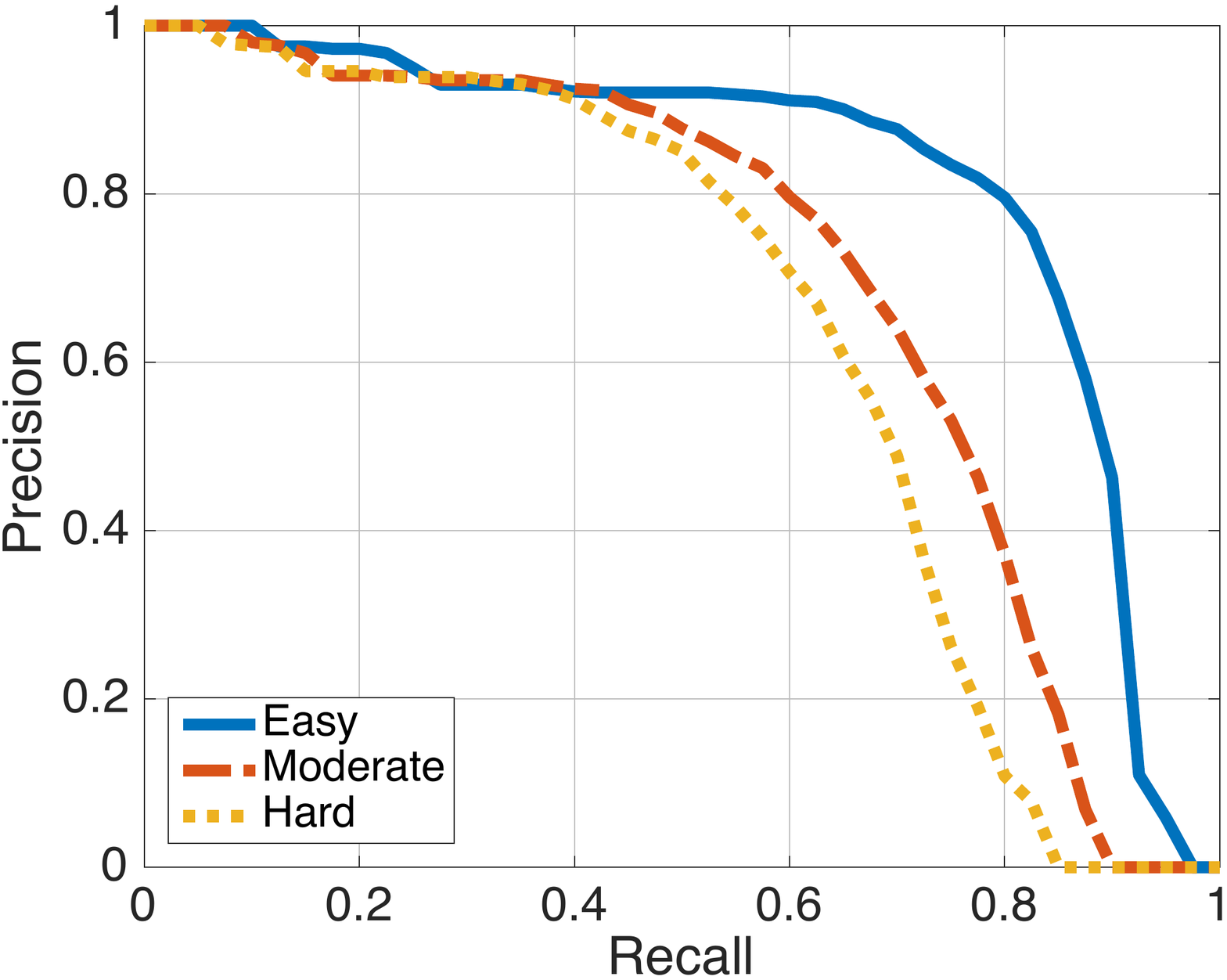}
    \label{fig:kitti_pr_cyc}}
}
\caption{Precision-Recall curves for the evaluation results on the KITTI \emph{test set}. ``Model B'' for cars and ``Model D'' for pedestrians and cyclists, all with eight filters in the hidden layers and trained without sparsity penalty, are used for the submission to the official test server. Best viewed in colour.}
\label{fig:kitti_pr}
\end{figure*}

\subsection{$\mathcal{L}_1$ Sparsity Penalty}
While the ReLU non-linearity helps to maintain sparsity in the intermediate representations, we propose to include an additional regulariser to incite the network to discard uninformative features and increase sparsity throughout the entire CNN stack.

The $\mathcal{L}_1$ loss has been shown to result in sparse representations with values being exactly zero \cite{murphy2012machine}, which is precisely the requirement for this model.
Whereas the sparsity of the output layer can be tuned with a detection threshold, we encourage sparsity in the intermediate layers by incorporating a penalty term using the $\mathcal{L}_1$ norm of each feature activation.

We normalise this $\mathcal{L}_1$ loss with respect to the spatial dimensions of the feature map in each layer.
This renders the influence of the sparsity penalty less dependent on the size of the input for a given parameter setting.

\section{EXPERIMENTS}
\label{sec:experiments}

\subsection{Dataset}
We use the well-known KITTI Vision Benchmark Suite \cite{Geiger2012CVPR} for training and evaluating our detection models.
The dataset consists of synchronised stereo camera and lidar frames recorded from a moving vehicle with annotations for eight different object classes, showing a wide variety of road scenes with different appearances.
We only use the 3D point cloud data to train and test the models.

There are 7,518 frames in the KITTI test set whose labels are not publicly available.
The labelled training data consist of 7,481 frames which we split into two sets for training and validation ($80\%$ and $20\%$ respectively). 
The object detection benchmark considers three classes for evaluation: cars, pedestrians and cyclists with 28,742; 4,487; and 1,627 training labels, respectively.

\subsection{Evaluation}
The benchmark evaluation on the official KITTI \emph{test set} is performed in 2D image space.
We therefore project our 3D detections into the 2D image plane using the provided calibration files and discard any detections that fall outside of the image.

The KITTI benchmark differentiates between easy, moderate and hard test categories depending on the bounding box size, object truncation and occlusion.
The hard test case considers the largest number of positives, whereas the most difficult examples are subsequently ignored for the moderate and easy test cases.
The official rankings are based on the average precision (AP) for the moderate cases.

After describing the training procedure, we present results for three experiments.
Firstly, we conduct a model comparison on the \emph{validation set} (Section \ref{sec:model_comparison}).
Secondly, based on the results of the model comparison, we select one model for each class and report results on the official KITTI \emph{test set} (Section \ref{sec:test_results}).
Lastly, we compare the timing results of models that were trained with and without the $\mathcal{L}_1$ sparsity penalty (Section \ref{sec:timing}).

\subsection{Training}
The networks are trained on 3D crops of positive and negative examples.
The number of positives and negatives is initially balanced with negatives being extracted randomly from the training data at locations that do not overlap with any of the positives.

In order to improve generalisation and to compensate for the fact that the input is discretised both spatially as well as in terms of angular resolution, the training data is augmented by translating the original front-facing positive training examples by a distance smaller than the size of the 3D grid cells and randomly rotating them by an angle that is smaller than the resolution of the angular bins.

Hard negative mining is performed every ten epochs by running the current model across the full point clouds in the training set.
In each round of hard negative mining, the ten highest scoring false positives per frame are added to the training set.

The filter weights are initialised as in \cite{he2015delving} and the networks are trained for 100 epochs with stochastic gradient descent with a momentum term of 0.9, a batchsize of 16, a constant learning rate of $10^{-3}$ and $\mathcal{L}_2$ weight decay of $10^{-4}$.
The model from the epoch with the highest AP on the \emph{validation set} is selected for the model comparison and the test submission.

For the timing experiments, we observed that selecting the models from the epoch with the highest AP on the validation set tends to favour models with a comparatively low sparsity in the intermediate representations.
Thus, the models after the full 100 epochs of training are used for the timing experiments to enable a fair comparison.

We implemented a custom C++ library for training and testing. For the largest models, training takes about three days on a cluster CPU node with 16 cores where each example in a batch is processed in a separate thread.

\subsection{Model Comparison}
\label{sec:model_comparison}

\begin{table*}
    \parbox[t]{\textwidth}{
    \centering
    \caption{AP in \% on the KITTI test set for methods only using point clouds (at the time of writing)}
    \label{tab:kitti_laser}
    \newcolumntype{Y}{>{\centering\arraybackslash}X}
    \begin{tabularx}{1.0\textwidth}{l r r c*{9}{Y}}
        \toprule
         & & & \multicolumn{3}{c}{Cars} & \multicolumn{3}{c}{Pedestrians} & \multicolumn{3}{c}{Cyclists} \\
        \cmidrule(r){4-6} \cmidrule(lr){7-9} \cmidrule(l){10-12}
        & Processor & Speed & Easy & Moderate & Hard & Easy & Moderate & Hard & Easy & Moderate & Hard \\
      \midrule
        \emph{Vote3Deep}				& 4-core 2.5GHz CPU & 1.1s & \textbf{76.79} & \textbf{68.24} & \textbf{63.23} & \textbf{68.39} & \textbf{55.37} & \textbf{52.59} & \textbf{79.92} & \textbf{67.88} & \textbf{62.98} \\
        Vote3D \cite{wang2015voting}    & 4-core 2.8GHz CPU & \textbf{0.5s} & 56.80 & 47.99 & 42.56 & 44.48 & 35.74 & 33.72 & 41.43 & 31.24 & 28.60 \\
        VeloFCN \cite{livehicle}        & 2.5GHz GPU        & 1.0s & 60.34 & 47.51 & 42.74 & - & - & - & - & - & - \\
        CSoR                            & 4-core 3.5GHz CPU & 3.5s & 34.79 & 26.13 & 22.69 & - & - & - & - & - & - \\
        mBoW \cite{behley2013laser}     & 1-core 2.5GHz CPU & 10s  & 36.02 & 23.76 & 18.44 & 44.28 & 31.37 & 30.62 & 28.00 & 21.62 & 20.93 \\
        \bottomrule
    \end{tabularx}
    }
\end{table*}

\begin{table*}
    \centering
    \caption{AP in \% on the KITTI test set for methods utilising both point clouds and images as indicated by *  (at the time of writing)}
    \label{tab:kitti_laser+vision}
    \newcolumntype{Y}{>{\centering\arraybackslash}X}
    \begin{tabularx}{1.0\textwidth}{l r r c*{9}{Y}}
        \toprule
         & & & \multicolumn{3}{c}{Cars} & \multicolumn{3}{c}{Pedestrians} & \multicolumn{3}{c}{Cyclists} \\
        \cmidrule(r){4-6} \cmidrule(lr){7-9} \cmidrule(l){10-12}
        & Processor & Speed & Easy & Moderate & Hard & Easy & Moderate & Hard & Easy & Moderate & Hard \\
        \midrule
        \emph{Vote3Deep}  & 4-core 2.5GHz CPU & \textbf{1.1s} & \textbf{76.79} & 68.24 & \textbf{63.23} & 68.39 & 55.37 & \textbf{52.59} & \textbf{79.92} & \textbf{67.88} & \textbf{62.98} \\
        \midrule
        MV-RGBD-RF* \cite{gonzalez2015multiview} & 4-core 2.5GHz CPU & 4s & 76.40 & \textbf{69.92} & 57.47 & \textbf{73.30} & \textbf{56.59} & 49.63 & 52.97 & 42.61 & 37.42 \\
        Fusion-DPM* \cite{premebida2014pedestrian} & 1-core 3.5GHz CPU & 30s & - & - & - & 59.51 & 46.67 & 42.05 & - & - & - \\
        \bottomrule
    \end{tabularx}
\end{table*}

Fast detection speeds are particularly important in the context of robotics.
As larger, more expressive models come at a higher computational cost and consequently run at slower speeds, this section investigates the trade-off between model capacity and detection performance on the \emph{validation set}.
Five architectures as summarised in Table \ref{tab:architectures} with up to three layers and different filter configurations are benchmarked against each other.
The ``Model D'' architecture is illustrated as an example in Figure \ref{fig:architecture_diagram}.

Small $3\times3\times3$ and $5\times5\times5$ kernels are used in the lower layers, followed by a ReLU non-linearity.
The architectures are designed so that the \emph{total} receptive field is slightly larger than the class-specific bounding boxes.
The network output is computed by a linear layer which is implemented as a convolutional filter whose kernel size gives the desired receptive field size for a given object class.

As can be seen in Fig. \ref{fig:model_comparison}, the non-linear, multi-layer networks clearly outperform the linear baseline, which is comparable to \cite{wang2015voting}.
First and foremost, this demonstrates that increasing the complexity and expressiveness of the models is extremely helpful for detecting objects in point clouds.

The resulting gains when increasing the number of convolutional filters in the hidden layers are moderate compared to the large improvement over the baseline which is achieved with only eight filters.
Similarly, increasing the receptive field of the filter kernels, while keeping the total receptive field of the networks the same, does not indicate a significant improvement in performance.

It is possible that these larger models are not sufficiently regularised.
Another potential explanation is that the easy interpretability of 3D data enables even relatively small models to capture most of the variation in the input representation which is informative for solving the task.

\subsection{Test Results}
\label{sec:test_results}

As the model comparison shows, increasing the number of filters or the kernel size does not significantly improve accuracy, while inevitably deteriorating the detection speed.
Consequently, we choose to limit ourselves to eight $3\times3\times3$ filters in each of the hidden layers for the test submission.

As the models can be run in parallel during deployment, they should ideally run at approximately the same detection speed.
Due to the larger physical size of cars, compared to pedestrians and cyclists, the corresponding networks need a larger filter kernel in the output layer to achieve the required total receptive field, having a negative effect on detection speed.
For the submission to the KITTI test server, we therefore select the ``Model B'' with two layers for cars, and the ``Model D'' with three layers for pedestrians and cyclists.
The PR curves of these models on the KITTI \emph{test set} are shown in Figure \ref{fig:kitti_pr}.

The performance of \emph{Vote3Deep} is compared against the other leading approaches for object detection in point clouds at the time of writing in Table \ref{tab:kitti_laser}.
\emph{Vote3Deep} establishes new state-of-the-art performance in this category for all three classes and all three difficulty levels.
The performance boost is particularly significant for cyclists with a margin of almost 40\% in the easy test case and more than doubling the AP in the other two test cases.

\emph{Vote3Deep} currently runs on CPU and is about two times slower than \cite{wang2015voting} and almost as fast as \cite{livehicle}, with the latter relying on GPU acceleration.
We expect that a GPU implementation of the sparse convolution layers will further improve the detection speed.

We also compare \emph{Vote3Deep} against methods that utilise both point cloud and image data at the time of writing in Table \ref{tab:kitti_laser+vision}.
Despite only using point cloud data, \emph{Vote3Deep} still performs better than these (\cite{gonzalez2015multiview,premebida2014pedestrian}) in the majority of test cases and only slightly worse in the remaining ones at a considerably faster detection speed.
For all three object classes, \emph{Vote3Deep} achieves the highest AP on the hard test cases, which considers the largest number of positive ground truth objects.

Overall, compared to the very deep networks used in vision (e.g. \cite{simonyan2014very, szegedy2015going, he2015deep}), these relatively shallow networks trained without any of the recently developed tricks are expressive enough to achieve significant performance gains.

Interestingly, cyclist detection benefits the most from the expressiveness of CNNs even though this class has the least number of training examples.
We conjecture that cyclists have a more distinctive shape in 3D compared to pedestrians and cars, which can be more easily confused with poles or vertical planes, respectively, and that \emph{Vote3Deep} models can exploit this complexity particularly well, despite the small number of positive training examples.

\subsection{Timing and Sparsity}
\label{sec:timing}

The three models from the test submission are also trained with different values for the $\mathcal{L}_1$ sparsity penalty to examine the effect of the penalty on detection speed and accuracy on the moderate test cases of the \emph{validation set} in Table \ref{tab:sparsity}.
The mean and standard deviation of the detection time per frame are measured on 200 frames.

Independent of whether the sparsity penalty is employed or not, pedestrians have the fastest detection speed as the receptive field of the networks is smaller compared to the other two classes.
The two-layer ``Model B'' for cars runs faster than the three-layer ``Model D'' for cyclists.

When imposing the $\mathcal{L}_1$ sparsity penalty during training, the detection speed at test time is improved by almost 40\% for cars at a negligible decrease in accuracy.
When applying a large penalty of $10^{-1}$, the activations of the pedestrian and cyclists models collapse to zero during training.
Yet, with a smaller penalty the detection speeds improve by about 15\%.

For the fastest cyclist model, the average precision decreases by 5\% compared to the baseline.
For pedestrians, however, we noted that the model without a penalty starts to overfit when training for the full 100 epochs.
In this case, the sparsity penalty helps to regularise the model and has a beneficial effect on the model's accuracy.

Notably, the sparsity penalty proves to be most useful for increasing the detection speed for cars where a larger penalty can be applied.
We conjecture that both the reduced number of intermediate layers as well as the larger receptive field help the model to learn significantly sparser, yet still highly informative, intermediate representations.

While the results clearly indicate that the L1 sparsity penalty has a beneficial effect on detection speed, a more rigorous investigation into the statistics of this gain would be useful, given the stochastic nature of the training algorithm. We leave this investigation for future work.

\begin{table}
    \parbox[b!]{\columnwidth}{
    \centering
    \caption{Detection speed in milliseconds and average precision for different values of the $\mathcal{L}_1$ sparsity penalty}
    \begin{tabularx}{\columnwidth}{l c c c c c c}
        \toprule
        & \multicolumn{2}{c}{Cars} & \multicolumn{2}{c}{Pedestrians} & \multicolumn{2}{c}{Cyclists} \\
        \cmidrule(r){2-3} \cmidrule(lr){4-5} \cmidrule(l){6-7}
        Penalty & Run-time & AP & Run-time & AP & Run-time & AP \\
        \midrule
        0           & 873$\pm$234                   & \textbf{0.76} & 508$\pm$119                   & 0.70          & 1055$\pm$301                  & \textbf{0.86} \\
        \midrule
        $10^{-3}$   & 819$\pm$211                   & 0.75          & 518$\pm$114                   & \textbf{0.73} & 1090$\pm$322                  & \textbf{0.86} \\
        $10^{-2}$   & 814$\pm$213                   & 0.74          & \textbf{426}$\pm$\textbf{84}  & 0.72          & \textbf{888}$\pm$\textbf{239} & 0.81 \\
        $10^{-1}$   & \textbf{553}$\pm$\textbf{134} & 0.75          & ---                           & ---           & ---                           & --- \\
        \bottomrule
    \end{tabularx}
    \label{tab:sparsity}}
\end{table}
\section{CONCLUSIONS}
\label{sec:conc}

This work performs object detection in point clouds at fast speeds with CNNs constructed from sparse convolutional layers based on the voting scheme introduced in \cite{wang2015voting}.
With the ability to learn hierarchical representations and non-linear decision boundaries, a new state of the art is established on the KITTI benchmark for detecting objects in point clouds.
\emph{Vote3Deep} also outperforms other methods that utilise information from both point clouds and images in most test cases.
Possible future directions include a more low-level input representation as well as a GPU implementation of the voting algorithm.

\balance

\section*{ACKNOWLEDGMENT}

The authors would like to acknowledge the support of this work by the EPSRC through grant number DFR01420, a Leadership Fellowship, a grant for Intelligent Workspace Acquisition, and a DTA Studentship; by Google through a studentship; and by the Advanced Research Computing services at the University of Oxford.
                                  
\bibliographystyle{bib_style/IEEEtran}
\bibliography{bib_style/IEEEabrv,Vote3Deep}

\begin{thebibliography}{10}
\providecommand{\url}[1]{#1}
\csname url@rmstyle\endcsname
\providecommand{\newblock}{\relax}
\providecommand{\bibinfo}[2]{#2}
\providecommand\BIBentrySTDinterwordspacing{\spaceskip=0pt\relax}
\providecommand\BIBentryALTinterwordstretchfactor{4}
\providecommand\BIBentryALTinterwordspacing{\spaceskip=\fontdimen2\font plus
\BIBentryALTinterwordstretchfactor\fontdimen3\font minus
  \fontdimen4\font\relax}
\providecommand\BIBforeignlanguage[2]{{%
\expandafter\ifx\csname l@#1\endcsname\relax
\typeout{** WARNING: IEEEtran.bst: No hyphenation pattern has been}%
\typeout{** loaded for the language `#1'. Using the pattern for}%
\typeout{** the default language instead.}%
\else
\language=\csname l@#1\endcsname
\fi
#2}}

\bibitem{krizhevsky2012imagenet}
A.~Krizhevsky, I.~Sutskever, and G.~E. Hinton, ``{ImageNet Classification with
  Deep Convolutional Neural Networks},'' \emph{Advances In Neural Information
  Processing Systems}, pp. 1--9, 2012.

\bibitem{simonyan2014very}
\BIBentryALTinterwordspacing
K.~Simonyan and A.~Zisserman, ``{Very deep convolutional networks for
  large-scale image recognition},'' \emph{ICLR}, pp. 1--14, 2015. [Online].
  Available: \url{http://arxiv.org/abs/1409.1556}
\BIBentrySTDinterwordspacing

\bibitem{szegedy2015going}
C.~Szegedy, W.~Liu, Y.~Jia, P.~Sermanet, S.~Reed, D.~Anguelov, D.~Erhan,
  V.~Vanhoucke, and A.~Rabinovich, ``{Going deeper with convolutions},'' in
  \emph{Proceedings of the IEEE Computer Society Conference on Computer Vision
  and Pattern Recognition}, vol. 07-12-June, 2015, pp. 1--9.

\bibitem{he2015deep}
\BIBentryALTinterwordspacing
K.~He, X.~Zhang, S.~Ren, and J.~Sun, ``{Deep Residual Learning for Image
  Recognition},'' \emph{arXiv preprint arXiv:1512.03385}, vol.~7, no.~3, pp.
  171--180, 2015. [Online]. Available:
  \url{http://arxiv.org/pdf/1512.03385v1.pdf}
\BIBentrySTDinterwordspacing

\bibitem{wang2015voting}
D.~Z. Wang and I.~Posner, ``{Voting for Voting in Online Point Cloud Object
  Detection},'' \emph{Robotics Science and Systems}, 2015.

\bibitem{Geiger2012CVPR}
A.~Geiger, P.~Lenz, and R.~Urtasun, ``{Are we ready for autonomous driving? the
  KITTI vision benchmark suite},'' in \emph{Proceedings of the IEEE Computer
  Society Conference on Computer Vision and Pattern Recognition}, 2012, pp.
  3354--3361.

\bibitem{livehicle}
\BIBentryALTinterwordspacing
B.~Li, T.~Zhang, and T.~Xia, ``{Vehicle Detection from 3D Lidar Using Fully
  Convolutional Network},'' \emph{arXiv preprint arXiv:1608.07916}, 2016.
  [Online]. Available: \url{https://arxiv.org/abs/1608.07916}
\BIBentrySTDinterwordspacing

\bibitem{maturana2015voxnet}
D.~Maturana and S.~Scherer, ``{VoxNet: A 3D Convolutional Neural Network for
  Real-Time Object Recognition},'' \emph{IROS}, pp. 922--928, 2015.

\bibitem{maturana20153d}
------, ``{3D Convolutional Neural Networks for Landing Zone Detection from
  LiDAR},'' \emph{International Conference on Robotics and Automation}, no.
  Figure 1, pp. 3471--3478, 2015.

\bibitem{graham2014spatially}
\BIBentryALTinterwordspacing
B.~Graham, ``{Spatially-sparse convolutional neural networks},'' \emph{arXiv
  Preprint arXiv:1409.6070}, pp. 1--13, 2014. [Online]. Available:
  \url{http://arxiv.org/abs/1409.6070}
\BIBentrySTDinterwordspacing

\bibitem{graham2015sparse}
\BIBentryALTinterwordspacing
------, ``{Sparse 3D convolutional neural networks},'' \emph{arXiv preprint
  arXiv:1505.02890}, pp. 1--10, 2015. [Online]. Available:
  \url{http://arxiv.org/abs/1505.02890}
\BIBentrySTDinterwordspacing

\bibitem{jampani2016learning}
V.~Jampani, M.~Kiefel, and P.~V. Gehler, ``{Learning Sparse High Dimensional
  Filters: Image Filtering, Dense CRFs and Bilateral Neural Networks},'' in
  \emph{IEEE Conf. on Computer Vision and Pattern Recognition (CVPR)}, 2016.

\bibitem{chen2016voxresnet}
\BIBentryALTinterwordspacing
H.~Chen, Q.~Dou, L.~Yu, and P.-A. Heng, ``{VoxResNet: Deep Voxelwise Residual
  Networks for Volumetric Brain Segmentation},'' \emph{arXiv preprint
  arXiv:1608.05895}, 2016. [Online]. Available:
  \url{http://arxiv.org/abs/1608.05895}
\BIBentrySTDinterwordspacing

\bibitem{dou2016automatic}
\BIBentryALTinterwordspacing
Q.~Dou, H.~Chen, L.~Yu, L.~Zhao, J.~Qin, D.~Wang, V.~C. Mok, L.~Shi, and P.~A.
  Heng, ``{Automatic Detection of Cerebral Microbleeds From MR Images via 3D
  Convolutional Neural Networks},'' \emph{IEEE Transactions on Medical
  Imaging}, vol.~35, no.~5, pp. 1182--1195, 2016. [Online]. Available:
  \url{http://ieeexplore.ieee.org}
\BIBentrySTDinterwordspacing

\bibitem{prasoon2013deep}
A.~Prasoon, K.~Petersen, C.~Igel, F.~Lauze, E.~Dam, and M.~Nielsen, ``{Deep
  feature learning for knee cartilage segmentation using a triplanar
  convolutional neural network},'' in \emph{Lecture Notes in Computer Science
  (including subseries Lecture Notes in Artificial Intelligence and Lecture
  Notes in Bioinformatics)}, vol. 8150 LNCS, no. PART 2, 2013, pp. 246--253.

\bibitem{glorot2011deep}
X.~Glorot, A.~Bordes, and Y.~Bengio, ``{Deep Sparse Rectifier Neural
  Networks},'' \emph{AISTATS}, vol.~15, pp. 315--323, 2011.

\bibitem{murphy2012machine}
K.~P. Murphy, \emph{{Machine Learning: A Probabilistic Perspective}}.\hskip 1em
  plus 0.5em minus 0.4em\relax MIT press, 2012, ch.~13, pp. 423--480.

\bibitem{he2015delving}
\BIBentryALTinterwordspacing
K.~He, X.~Zhang, S.~Ren, and J.~Sun, ``{Delving Deep into Rectifiers:
  Surpassing Human-Level Performance on ImageNet Classification},'' \emph{arXiv
  preprint arXiv:1502.01852}, pp. 1--11, 2015. [Online]. Available:
  \url{https://arxiv.org/abs/1502.01852}
\BIBentrySTDinterwordspacing

\bibitem{behley2013laser}
J.~Behley, V.~Steinhage, and A.~B. Cremers, ``{Laser-based segment
  classification using a mixture of bag-of-words},'' in \emph{IEEE
  International Conference on Intelligent Robots and Systems}, 2013, pp.
  4195--4200.

\bibitem{gonzalez2015multiview}
A.~Gonzalez, G.~Villalonga, J.~Xu, D.~Vazquez, J.~Amores, and A.~M. Lopez,
  ``{Multiview random forest of local experts combining RGB and LIDAR data for
  pedestrian detection},'' in \emph{IEEE Intelligent Vehicles Symposium,
  Proceedings}, vol. 2015-Augus, 2015, pp. 356--361.

\bibitem{premebida2014pedestrian}
C.~Premebida, J.~Carreira, J.~Batista, and U.~Nunes, ``{Pedestrian detection
  combining RGB and dense LIDAR data},'' in \emph{IEEE International Conference
  on Intelligent Robots and Systems}, 2014, pp. 4112--4117.

\end{thebibliography}

\end{document}